\documentclass[conference]{IEEEtran}
\IEEEoverridecommandlockouts
\usepackage{cite}
\usepackage{amsmath,amssymb,amsfonts}
\usepackage{algorithmic}
\usepackage{graphicx}
\usepackage{textcomp}
\usepackage{xcolor}
\usepackage{subcaption}

\def\BibTeX{{\rm B\kern-.05em{\sc i\kern-.025em b}\kern-.08em
    T\kern-.1667em\lower.7ex\hbox{E}\kern-.125emX}}
\begin{document}

\title{Mono vs Multilingual BERT for Hate Speech Detection and Text Classification: A Case Study in Marathi\\
\thanks{Supported by L3Cube Pune}
}

\author{\IEEEauthorblockN{Abhishek Velankar}
\IEEEauthorblockA{
% \textit{Information Technology} \\
\textit{Pune Institute of Computer Technology}\\
\textit{L3Cube Pune}\\
% Pune, India \\
velankarabhishek@gmail.com}
\and
\IEEEauthorblockN{Hrushikesh Patil}
\IEEEauthorblockA{
% \textit{Information Technology} \\
\textit{Pune Institute of Computer Technology}\\
\textit{L3Cube Pune}\\
% Pune, India \\
hrushi2900@gmail.com}
\and

\IEEEauthorblockN{Raviraj Joshi}
\IEEEauthorblockA{
% \textit{Computer Science} \\
\textit{Indian Institute of Technology Madras}\\
\textit{L3Cube Pune}\\
% Chennai, India \\
ravirajoshi@gmail.com}
% \and
% \and
}

\maketitle

\begin{abstract}
Transformers are the most eminent architectures used for a vast range of Natural Language Processing tasks. These models are pre-trained over a large text corpus and are meant to serve state-of-the-art results over tasks like text classification. In this work, we conduct a comparative study between monolingual and multilingual BERT models. We focus on the Marathi language and evaluate the models on the datasets for hate speech detection, sentiment analysis, and simple text classification in Marathi. We use standard multilingual models such as mBERT, indicBERT, and xlm-RoBERTa and compare them with MahaBERT, MahaALBERT, and MahaRoBERTa, the monolingual models for Marathi. We further show that Marathi monolingual models outperform the multilingual BERT variants in five different downstream fine-tuning experiments. We also evaluate sentence embeddings from these models by freezing the BERT encoder layers. We show that monolingual MahaBERT-based models provide rich representations as compared to sentence embeddings from multi-lingual counterparts. However, we observe that these embeddings are not generic enough and do not work well on out-of-domain social media datasets. We consider two Marathi hate speech datasets L3Cube-MahaHate, HASOC-2021, a Marathi sentiment classification dataset L3Cube-MahaSent, and Marathi Headline, Articles classification datasets.
\end{abstract}


\section{Introduction}
The language models like BERT, built over the transformer architecture, have gained a lot of popularity due to the promising results on an extensive range of natural language processing tasks. These large models make use of attention mechanism from transformers and understand the language deeper in terms of context. These models can be fine-tuned on domain-specific data to obtain state-of-the-art solutions. 

More recently, there has been a significant amount of research on monolingual and multilingual language models, specifically the BERT variants. Due to the variety of text corpus in terms of languages used for training, multilingual models find notable benefits over multiple applications, specifically for the languages that are low in resources \cite{DBLP:conf/acl/PiresSG19,pires-etal-2019-multilingual,joshi2021evaluation}. However, the monolingual models, when used in the corresponding language, outperform the multilingual versions in tasks like text classification. Both the categories of models find their use in several problems like next sentence prediction, named entity recognition, sentiment analysis, etc. Recently, a substantial amount of work can be seen with the use of these models in native languages. \cite{DBLP:journals/corr/abs-2112-04329} propose monolingual BERT models for the Arabic language and show that these models achieve state-of-the-art performances. Additionally, \cite{le-etal-2020-flaubert-unsupervised}, \cite{DBLP:journals/corr/abs-2108-13741}, \cite{DBLP:journals/corr/abs-2112-10553} show that the single language models, when used for the corresponding language tasks, perform more efficiently than the multilingual variants. \cite{DBLP:journals/corr/abs-1910-03806} analyze the effectiveness of multilingual models over the monolingual counterparts for 6 different languages including English and German. Our work focuses on hate speech detection, sentiment analysis, and simple text classification in Marathi \cite{DBLP:journals/corr/abs-2110-12200,DBLP:conf/wassa/KulkarniMLKJ21,kulkarni2022experimental,velankar2022l3cube}. We evaluate monolingual and multilingual BERT models on Marathi corpus to compare the performance. A similar analysis for Hindi and Marathi named entity recognition has been performed in \cite{litake2022mono}.

Marathi is a regional language in India. It is majorly spoken by the people in Maharashtra \cite{DBLP:journals/corr/abs-2202-01159}. Additionally, after Hindi and Bengali, it is considered as the third most popular language in India \cite{DBLP:conf/ihci/JoshiGJ19,DBLP:journals/corr/IslamJA17}. However, the Marathi language is greatly overlooked in terms of language resources which suggests the need of widening the research in this area.

In this work, we perform a comparative analysis of monolingual and multilingual BERT models for Marathi. We fine-tune these models over the Marathi corpus, which contains hate speech detection and simple text classification datasets. We consider standard multilingual models i.e mBERT, indicBERT, and xlm-RoBERTa, and compare them with Marathi monolingual counterparts i.e. MahaBERT, MahaALBERT, and MahaRoBERTa. We further show that the monolingual models when used in Marathi, outperform the multilingual equivalents. Moreover, we evaluate sentence representations from these models and show that the monolingual models provide superior sentence representations. The advantage of using monolingual models is more visible when extracted sentence embeddings are used for classification. This research is aimed to help the community by giving an insight into the appropriate use of these single and multilingual models when applied to single language tasks.

\section{Related work}

The BERT is currently one of the most effective language models in terms of performance when different NLP tasks like text classification are concerned. The previous research has shown how BERT captures the language context in an efficient way \cite{jawahar-etal-2019-bert}, \cite{tenney-etal-2019-bert}, \cite{de-vries-etal-2020-whats}.

Recently, a lot of work can be seen in single and multi-language NLP applications. Several efforts have been made to build monolingual variants of BERT and shown to be effective over a quantity of single language downstream tasks. In \cite{DBLP:journals/corr/abs-2012-02110} authors publish a German monolingual BERT model based on  RoBERTa. The experiments have been performed in the tasks like named entity recognition (NER) and text classification to evaluate the model performance. They further propose that, with the only little tuning of hyperparameters, the model outperformed all other tested German and multilingual BERT models. A monolingual RoBERTa language model trained on Czech data has been presented in \cite{DBLP:journals/corr/abs-2105-11314}. The authors show that the model significantly outperforms equally-sized multilingual and Czech language-oriented model variants. Other works for single language-specific BERT models include models built in Vietnamese, Hindi, Bengali, etc. \cite{DBLP:journals/corr/abs-2003-00744}, \cite{DBLP:journals/corr/abs-2011-02323}. In \cite{DBLP:journals/corr/abs-2111-04530} authors propose model evaluations on toxicity detection in Spanish comments. They show that transformers obtain better results than statistical models. Furthermore,  they conclude monolingual BERT models provide better results in their pre-trained language as compared to multilingual models.

\section{Datasets}

\begin{itemize}
\item \textbf{HASOC’21 Marathi dataset \cite{modha2021overview}:}\\
A Marathi binary dataset provided in HASOC’21 shared task divided into hateful and non-hateful categories. It consists of a total of 1874 training and 625 testing samples.
\\ 
\item \textbf{L3Cube-MahaHate \cite{velankar2022l3cube}:}\\
A hate speech detection dataset in Marathi consists of 25000 tweet samples divided into 4 major classes namely hate, offensive, profane, and not. The dataset consists of 21500 train, 2000 test, and 1500 validation examples.
\\
\item \textbf{Articles:}\\
A text classification dataset containing Marathi news articles classified into sports, entertainment, and lifestyle with 3823 train, 479 test, and 477 validation samples.
\\ 
\item \textbf{Headlines:}\\
A Marathi news headlines dataset containing the headlines containing three classes viz. entertainment, sports, and state. It consists of 9672 train, 1210 test, and 1210 validation samples.
\\
\item \textbf{L3Cube-MahaSent \cite{DBLP:conf/wassa/KulkarniMLKJ21}:}\\
A Sentiment Analysis dataset in Marathi
includes tweets classified as positive,
negative, and neutral. It has 12114 train, 2250 test, and 1500 validation examples.

\end{itemize}
\section{Experiments}
\subsection{Transformer models}
BERT is a deep transformer model, pre-trained over large text corpus in a self-supervised manner, and provides a great ability to promptly adapt to a broad range of downstream tasks. There are a lot of different flavors of BERT available openly, some popular variants are ALBERT and RoBERTa. In this work, we are focusing on both multilingual and monolingual models for text classification and hate speech detection tasks. Following standard multilingual BERT models which use Marathi as one of the training languages are used:

\begin{itemize}
\item \textbf{Multilingual-BERT (mBERT)\footnote{https://huggingface.co/bert-base-multilingual-cased}: }
It is a BERT-base model \cite{DBLP:journals/corr/abs-1810-04805} trained on and usable with 104 languages with Wikipedia using a masked language modeling (MLM) and next sentence prediction (NSP) objective.
\\ 
\item \textbf{IndicBERT\footnote{https://huggingface.co/ai4bharat/indic-bert}: }
a multilingual ALBERT model released by Ai4Bharat \cite{kakwani2020indicnlpsuite}, trained on large-scale corpora. The training languages include 12 major Indian languages. The model has been proven to be working better for the tasks in Indic languages. 
\\
\item \textbf{XLM-RoBERTa\footnote{https://huggingface.co/xlm-roberta-base}: }
a multilingual version of the RoBERTa model \cite{DBLP:journals/corr/abs-1911-02116}. It is pre-trained on 2.5TB of filtered CommonCrawl data containing 100 languages with the Masked language modeling (MLM) objective and can be used for downstream tasks.
\end{itemize} 

To compare with the above models, the following Marathi monolingual models are used \cite{DBLP:journals/corr/abs-2202-01159}:
\begin{itemize}
\item \textbf{MahaBERT\footnote{https://huggingface.co/l3cube-pune/marathi-bert}: }
a multilingual BERT model fine-tuned on L3Cube-MahaCorpus and other publicly available Marathi monolingual datasets containing a total of 752M tokens.
\\
\item \textbf{MahaAlBERT\footnote{https://huggingface.co/l3cube-pune/marathi-albert}: }
It is a Marathi monolingual model extended from AlBERT, trained on L3Cube-MahaCorpus and other publicly available Marathi monolingual datasets.
\\
\item \textbf{MahaRoBERTa\footnote{https://huggingface.co/l3cube-pune/marathi-roberta}: }
It is a Marathi RoBERTa model built upon a multilingual RoBERTa (xlm-roberta-base) model fine-tuned on L3Cube-MahaCorpus and other publicly available Marathi monolingual datasets.
\end{itemize}

\subsection{Evaluation results} 
\begin{table*}

\caption{Classification accuracies for monolingual and multilingual models}\label{tab1}
\setlength\tabcolsep{2pt}
\renewcommand{\arraystretch}{1.4}
\centering
\begin{tabular}{|c|c|c|c|c|c|c|}
\hline
\multicolumn{1}{|p{2cm}|}{\centering \textbf{Model}} &  \multicolumn{1}{|p{2cm}|}{\centering \textbf{Training} \\ \textbf{Mode}}  &  \multicolumn{1}{|p{1.5cm}|}{\centering \textbf{HASOC}} & \multicolumn{1}{|p{1.5cm}|}{\centering \textbf{L3Cube-} \\ \textbf{MahaHate}} &  \multicolumn{1}{|p{1.5cm}|}{\centering \textbf{L3Cube-} \\ \textbf{MahaSent}}  & \multicolumn{1}{|p{1.5cm}|}{\centering \textbf{Articles}}  & \multicolumn{1}{|p{1.5cm}|}{\centering \textbf{Headlines}}   \\
\hline
\multicolumn{7}{|c|}{\textbf{Multilingual BERT Variants}} \\
\cline{1-7} 
mBERT &  Freeze  & 0.770 & 0.516 &  0.653  & 0.901 & 0.907  \\
 &  Non-Freeze  & \textbf{0.875} & 0.783 &  0.786  & 0.976 & \textbf{0.947}  \\
 \hline
IndicBERT &  Freeze  & 0.710 & 0.436 &  0.656  & 0.828 & 0.877  \\
 &  Non-Freeze  & 0.870 & 0.711 &  \textbf{0.833}  & \textbf{0.987} & 0.937 \\
 \hline
xlm-RoBERTa &  Freeze  & 0.755 & 0.487 &  0.666  & 0.91 & 0.79  \\
 &  Non-Freeze  & 0.862 & \textbf{0.787} &  0.820 & 0.985 & 0.925  \\
\hline
\multicolumn{7}{|c|}{\textbf{Monolingual BERT Variants}} \\
\cline{1-7} 
MahaBERT &  Freeze  & 0.824 & 0.580 &  0.666  & 0.939 & 0.907  \\
 &  Non-Freeze  & 0.883 & 0.802 &  0.828  & 0.987 & 0.944  \\
 \hline
MahaAlBERT &  Freeze  & 0.800 & 0.587 & 0.717  & 0.991 & 0.927  \\
 &  Non-Freeze  & 0.866 & 0.764 &  \textbf{0.841}  & \textbf{0.991} & \textbf{0.945} \\
 \hline
MahaRoBERTa &  Freeze  & 0.782 & 0.531 & 0.698  & 0.904 & 0.864  \\
 &  Non-Freeze  & \textbf{0.890} & \textbf{0.803} &  0.834 & 0.985 & 0.942  \\
\hline
\end{tabular}
\end{table*}

\begin{figure*}[hbt!]
\centering
\begin{subfigure}[b]{0.5\textwidth}
\includegraphics[width=\textwidth]{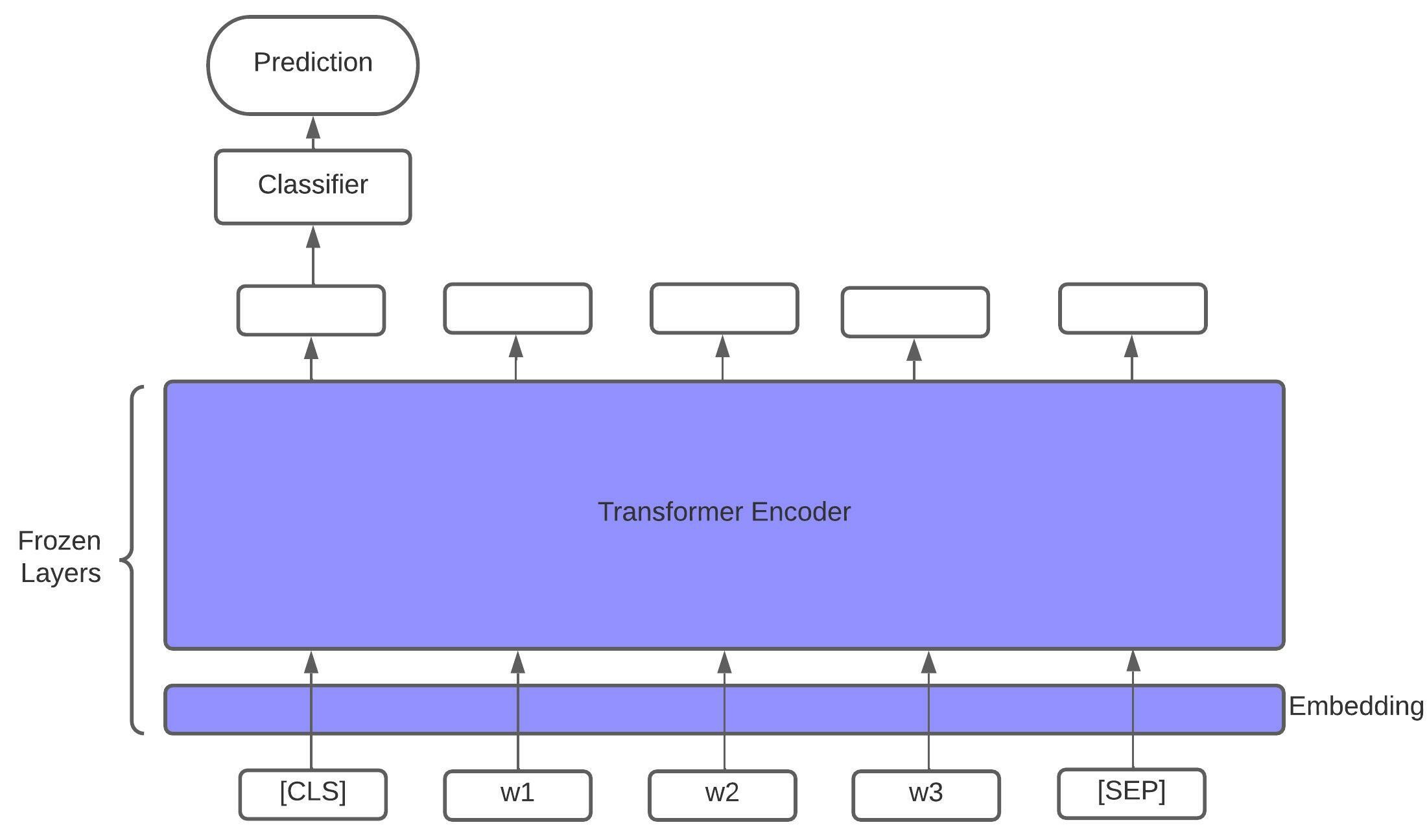}
% \caption{2-class classification}
\label{fig:conf2}
\end{subfigure}
\begin{subfigure}[b]{0.4\textwidth}
\includegraphics[width=\textwidth]{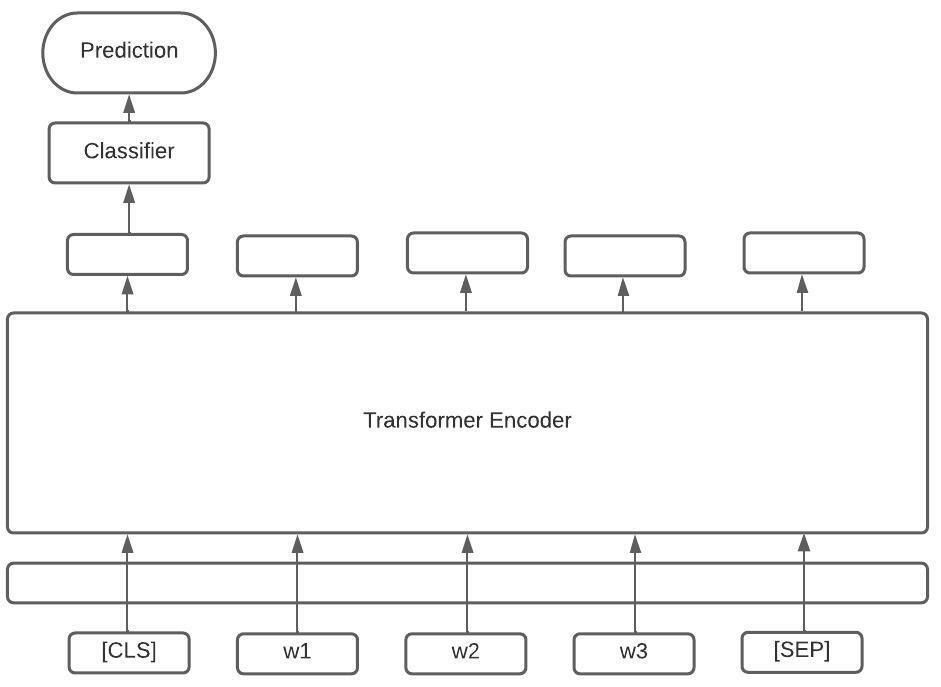}
% \caption{4-class classification}
\label{fig:conf4}
\end{subfigure}
\caption{BERT architectures with freeze and non-freeze training mode}
\end{figure*}
The BERT transformer models have been evaluated on hate speech detection and text classification datasets. We used standard multilingual BERT variants namely indicBERT, mBERT and xlm-RoBERTa to obtain baseline classification results. Additionally, monolingual Marathi models have been used for comparison. These single language models include MahaBERT, MahaAlBERT, and MahaRoBERTa are based on the  BERT-base, AlBERT, and RoBERTa-base models respectively.

The experiments have been performed in two schemes. Firstly, we obtained the results by fine-tuning all the BERT layers i.e pre-trained layers as well as classification layers. Furthermore, we froze the pre-trained embedding and encoder layers and trained only the classifier to obtain the results. Using this setup we aim to evaluate the sentence embeddings generated by these monolingual and multilingual models. All the classification results are displayed in Table \ref{tab1}.

% \begin{table}[htbp]
% \caption{Table Type Styles}
% \begin{center}
% \begin{tabular}{|c|c|c|c|}
% \hline
% \textbf{Table}&\multicolumn{3}{|c|}{\textbf{Table Column Head}} \\
% \cline{2-4} 
% \textbf{Head} & \textbf{\textit{Table column subhead}}& \textbf{\textit{Subhead}}& \textbf{\textit{Subhead}} \\
% \hline
% copy& More table copy$^{\mathrm{a}}$& &  \\
% \hline
% \multicolumn{4}{l}{$^{\mathrm{a}}$Sample of a Table footnote.}
% \end{tabular}
% \label{tab1}
% \end{center}
% \end{table}

% \begin{figure}[htbp]
% \centerline{\includegraphics{fig1.png}}
% \caption{Example of a figure caption.}
% \label{fig}
% \end{figure}

For all the monolingual and multilingual models, the frozen settings i.e freezing BERT embedding and encoder layers are underperforming as compared to their non-freeze counterparts. The difference in accuracy is too high for L3Cube-MahaSent and L3Cube-MahaHate. This indicates that the pre-trained models do not provide generic discriminative sentence embeddings for the classification task. However, the mono-lingual model does provide better sentence embeddings as compared to the multi-lingual counterpart. This shows the importance of monolingual pretraining for obtaining rich sentence embeddings. Since the pre-training data is mostly comprised of Marathi news articles the frozen setting works comparatively well on the Articles and Headlines dataset.  In general, the monolingual models have outperformed the multilingual models on all the datasets. For hate speech detection datasets, particularly the MahaRoBERTa model is working the best. In the case of other text classification datasets, the MahaAlBERT model is giving the best accuracy.

\section{Conclusion}
In this paper, we have presented a comparison between monolingual and multilingual transformer-based models, particularly the variants of BERT.  We have evaluated these models on hate speech detection and text classification datasets. We have used standard multilingual models namely mBERT, indicBERT, and xlm-RoBERTa for evaluation. On the other hand, we have used Marathi monolingual models trained exclusively on large Marathi corpus i.e. MahaBERT, MahaAlBERT, and MahaRoBERTa for comparison. The MahaAlBERT model performs the best in the case of simple text classification whereas MahaRoBERTa gives the best results for hate speech detection tasks. The monolingual versions for all the datasets have outperformed the standard multilingual models when focused on single language tasks. The monolingual models also provide better sentence representations. However, these sentence representations do not generalize well across the tasks, thus highlighting the need for better sentence embedding models.

\section*{Acknowledgment}

This work was done under the L3Cube Pune mentorship program. We would like to express our gratitude towards our mentors at L3Cube for their continuous support and encouragement.

\bibliographystyle{IEEEtran}
\bibliography{main}
\end{document}